\documentclass[a4paper]{article}

\usepackage{INTERSPEECH2021}
\usepackage{subcaption}
\usepackage{hyperref}
\usepackage[mathscr]{euscript}
\usepackage[ruled,vlined]{algorithm2e}

\title{Continual Speaker Adaptation for Text-to-Speech Synthesis}
\name{Hamed Hemati$^1$, Damian Borth$^1$}
\address{
  $^1$University of St.Gallen, Switzerland}
\email{hamed.hemati@unisg.ch, damian.borth@unisg.ch}

\begin{document}

\maketitle
\begin{abstract}
Training a multi-speaker Text-to-Speech (TTS) model from scratch is computationally expensive and adding new speakers to the dataset requires the model to be re-trained. The naive solution of sequential fine-tuning of a model for new speakers can lead to poor performance of older speakers. This phenomenon is known as catastrophic forgetting. In this paper, we look at TTS modeling from a continual learning perspective, where the goal is to add new speakers without forgetting previous speakers. Therefore, we first propose an experimental setup and show that serial fine-tuning for new speakers can cause the forgetting of the earlier speakers. Then we exploit two well-known techniques for continual learning, namely experience replay and weight regularization. We reveal how one can mitigate the effect of degradation in speech synthesis diversity in sequential training of new speakers using these methods. Finally, we present a simple extension to experience replay to improve the results in extreme setups where we have access to very small buffers.
   
\end{abstract}

\section{Introduction}

Catastrophic Forgetting (CF) is a well-known problem in neural networks \cite{FRENCH1999128, Goodfellow-et-al-2016} and has been studied for many years. Different continual learning (also called lifelong learning) approaches tackle this problem with various points of view, such as regularizing the network's weights and rehearsal practices. For example, in \cite{Kirkpatrick3521} the authors propose Elastic Weight Consolidation (EWC) as a regularization term in the objective function to prevent the model from moving too much away from the previous weight state based on the importance of each module in the network. In \cite{Rebuffi_2017_CVPR} they propose an approach based on replaying examples from past tasks kept in a buffer alongside the model when training the model on new tasks. 

Continual learning in TTS models has various advantages, such as extending an existing multi-speaker TTS system, reducing training costs, and improving the speech quality of an existing speaker with new incoming data, to just name a few. Despite the extensive study of continual learning for domains like image and text, there has been little work in the speech domain to overcome CF. Yet, the focus in the speech domain has been mainly on automatic speech recognition \cite{Sadhu2020, xue2019multitask}. For TTS, previous works have concentrated on transfer learning and meta-learning methods for adaptation of new speakers \cite{jia2018transfer, chen2018sample}. These methods are usually trained with very large datasets consisting of high variant speech characteristics, and the hope is that a new speaker's vocal characteristics would be close to one of the speakers in the pre-trained model.

\begin{figure}[t]
  \centering
  \includegraphics[width=0.9\linewidth]{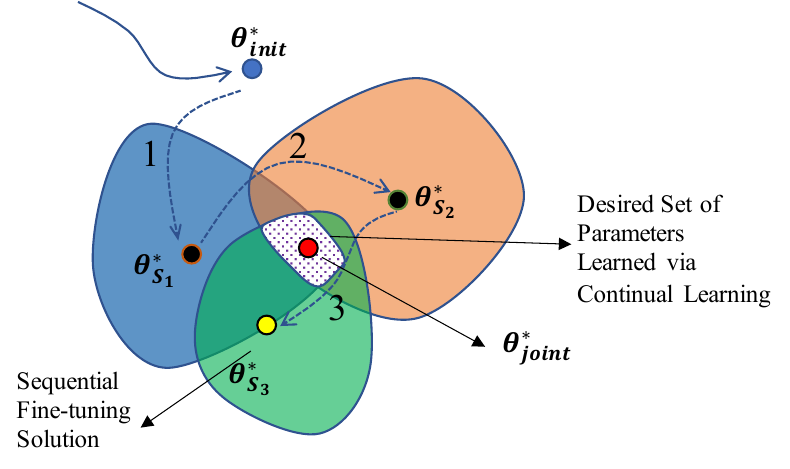}
  \caption{Illustration of the difference between optimal solutions for sequential fine-tuning (yellow point) vs. good possible solutions with proper continual learning (white dotted area) in a 2D parameter space $\theta$. The blue, orange, and green regions correspond to the low-loss areas for speakers 1, 2, and 3 in the sequence correspondingly.}
  \label{fig:solution_space}
\end{figure}

With the success of recent neural TTS models, \cite{46150, shen2018natural, ping2018deep}, the desire for continually adding new speakers to an existing TTS model and proper methods for measuring forgetting, backward and forward transfer, etc. becomes more important. To evaluate continual learning for TTS, we design a framework to enable this continual behavior and measurement. By considering the parameters of a joint-trained model $\theta_{joint}^*$ on all seen speakers as the best possible solution for a TTS model, we try to find a set of parameters $\theta$ that performs close to $\theta_{joint}^*$ as shown in Figure \ref{fig:solution_space}. Our contributions in this paper are as follows:

\begin{itemize}
    \item We propose a framework for the continual learning of speaker adaptation.
    \item We benchmark two popular CL methods with additional insights on results.
    \item  We extend experience replay for an extreme case where we have access to a very small buffer.
\end{itemize}

\section{Experimental Setup}

In this section, we introduce the proposed framework and explain the details of the setups and datasets that are used throughout this paper.

\subsection{Framework}

Our framework consists of three major components, namely the Base TTS Model (Base-TTS), Speaker Encoder (SE), and Speaker Verifier (SV). For the Base-TTS module, it is essential to choose a model that works appropriately for standard multi-speaker datasets. Therefore we adopt Tacotron 2 \cite{shen2018natural} and build upon an open-source implementation of it provided by Nvidia\footnote{\url{https://github.com/NVIDIA/DeepLearningExamples}}. One can, however, use a different model without the need to change the other parts of the framework. The Mel-spectrograms generated by Base-TTS are converted to waveforms with a neural vocoder; in our experiments, we use WaveRNN \cite{kalchbrenner2018efficient}.

\begin{figure}[t]
  \centering
\includegraphics[width=\linewidth]{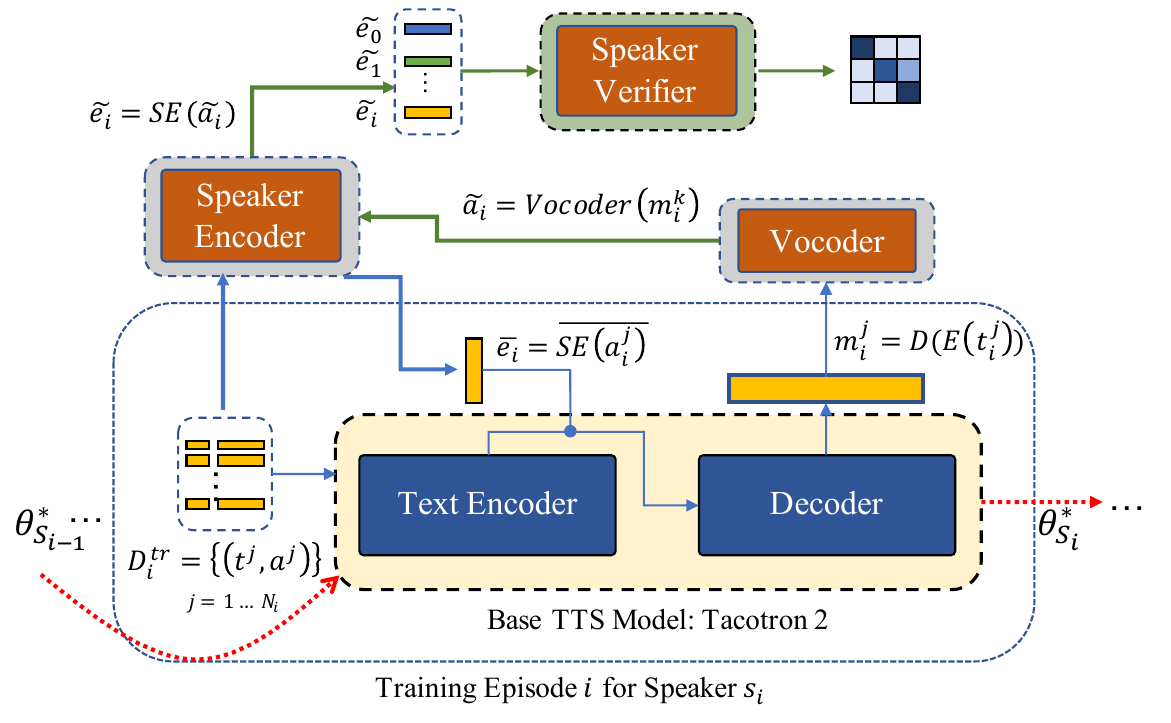}
  \caption{Schematic of the proposed continual learning setup for episode $s_i$. Modules inside gray boxes are pre-trained and frozen during training. The modules inside the green and yellow boxes are trained before and during each episode correspondingly. }
  \label{fig:framework}
\end{figure}

In the SE module, similar to \cite{jia2018transfer}, we employ a pre-trained speaker encoder based on \cite{wan2018generalized}. This module plays a significant role in the overall performance of the framework. We exploit a released checkpoint by Mozilla TTS \footnote{\url{https://github.com/mozilla/TTS/wiki/Released-Models}} trained on a combination of multi-speaker English datasets. The SV module directly works with speech embeddings obtained via SE, therefore we found a linear classifier works well in practice. In Figure \ref{fig:framework} we demonstrate the different modules and the interactions between them.

\subsection{Datasets}
We run our experiments in two languages, English and German. For English, we adopt VCTK \cite{yamagishi2019vctk}, and for German, we use the CommonVoice-DE (CVDE) split provided by \cite{Nekvinda2020}. Due to the high number of parameters in the base TTS model and the low number of utterances for each speaker, fine-tuning the first speaker in the sequence of speakers for a randomly initialized model would be very unstable, and it might not converge. Hence we divide the dataset into two splits. We use one split as a pre-training split (PR-Split) to pre-train the Base-TTS module to get $\theta_{init}^*$ and use a continual learning split (CL-Split) to run the experiments. To further improve the stability of initialization $\theta_{init}^*$  in our low-data setting, we found mixing the PR splits with a single speaker dataset beneficial without any possibility of overlapping with the future speakers. We exploited LJSpeech \cite{ljspeech17} for English, and CSS10-DE \cite{park2019css10} for German, as our single-speaker datasets. In Table \ref{tab:datasets} we show the details of the datasets used.

\begin{table}[th]
  \caption{Dataset Split Details}
  \label{tab:datasets}
  \centering
  \begin{tabular}{clll}
    \toprule
    \textbf{Dataset}      & \textbf{\# of Speakers} &  \textbf{PR-Split}     & \textbf{CL-Split}           \\
    \midrule
    VCTK & 109 & 89 & 20   \\
    LJSpeech & 1 & 1 & -   \\
    \hline
    CVDE & 39 & 24 & 15   \\
    CSS10-DE & 1 & 1 & -   \\
    \bottomrule
  \end{tabular}
\end{table}

\subsection{Performance Measurement}

Since the focus of the paper is on the continual learning perspective of TTS models, we assume that our model can converge for fine-tuning to a new speaker with the data provided for each speaker in the CL-Split of each dataset. This is possible with a proper choice of Base-TTS module and a good initialization $\theta_{init}^*$ provided via pre-training.

In our experiments, we observed that the model could generate good results for every new speaker regardless of its performance on the previously seen speakers. As the number of speakers increases over time, crow sourcing ways of measuring the speaker identities become more time-consuming and expensive. Therefore, we propose to instead use the accuracy of synthesized speech utterances in speaker verification using the SV module. 

At the beginning of each episode of training, we train the SV module with ground truth data $\{(e^j, c); j=1...N\} \forall c \in \{1..M\}$ where $e$ and $c$ are the speech embeddings of the ground truth (GT) audio files and speaker labels, and $N$ and $M$ are the numbers of audio files for each speaker and number of speakers seen so far respectively. To avoid distributional shift between synthesized speech files and GT speech files, we first reconstruct GT speech waveforms with the Vocoder. Training SV even for a sequence of 20 speakers is quick and always reaches an accuracy above 99\% after only 15 epochs which takes less than a minute on a CPU. 

After each episode is finished, we synthesize 5 utterances per speaker, and compute their embeddings $\{{\Tilde{e_i}}^j; 1 \leq i \leq M, 1 \leq j \leq 5 \}$. Similar to \cite{chaudhry2018efficient}, we calculate Retained Accuracy (RA) and Forgetting (FG) as a measure of diversity in the synthesized speech over time as below:

\newcommand{\RA}{\operatorname{RA}}
\newcommand{\FG}{\operatorname{FG}}

\begin{equation}
    \RA_{T} = \frac{1}{T} \sum_{i=1}^{T} a_{T,i} 
\end{equation}

\begin{equation}
    \FG_{T} = \frac{1}{T-1} \sum_{i-1}^{T-1} \max_{l \in 1,..,T-1} (a_{l,i} - a_{T,i})
\end{equation}

where $a_{i,l}$ is accuracy of speaker $i$ at episode $l$ and $T$ is the indicator of the current episode.

\section{Methods}

This section explains the details of various methods that we use in our experiments. In our setup, each speaker is considered a separate task. The average speech embeddings of the training data of each speaker ${\Bar{e}}_i$ is used as a task indicator to distinguish between different speakers for both training and inference. Throughout all experiments, we assume that we have a set of dataset pairs $\mathcal{D}_{CL}=\{(\mathcal{D}_{i}^{tr}, \mathcal{D}_{i}^{ev}); 1 \leq i \leq N \}$ where each pair represents train and evaluation sets for one distinct speaker, and $M$ is the total number of speakers. 

\subsection{Joint Training}
In Joint Training (JT), we train the model with data from all speakers together. Accordingly, we directly use $\mathcal{D}_{CL}$ for training and evaluation. Here have only one phase of training. JT is considered as the upper bound of the model's performance in the speaker verification accuracy due to the reason that it has access to all speakers at once. The objective for JT is as follows:

\begin{equation}
    \theta_{joint}^{*} = \operatorname*{arg\,min}_\theta {\sum_{t=1}^{t=M}{\mathscr{L}_{TTS}(D_{tr}^t|\theta = \theta_{init}^*})}
\end{equation}

\subsection{Continual Learning}
In the CL-based methods, since we sequentially train the models for new speakers, we call each complete training phase of a model till convergence an episode; hence we have $M$ episode in total. It is  important to reset the optimizer's state for momentum and adaptive learning rate-based gradient descent algorithm to ensure that a new speaker's optimization process does not use information from past gradients. In each method, we explain how the optimal parameter for every episode is obtained.

\subsubsection{Sequential Adaptation (SA)}
In Sequential, Adaptation, we fine-tune the model with the weights provided from the previous episodes and aim at reducing evaluation loss for the current task (speaker):

\begin{equation}
    \theta_{t}^{*} = \operatorname*{arg\,min}_\theta {\mathscr{L}_{TTS}(D_{tr}^t|\theta=\theta_{t-1}^*)}, \forall t \in \{1,...,M\}
\end{equation}

We consider SA as the lower bound for CL comparison in our setup as it completely ignores previous speakers and only optimizes for the new task.

\subsubsection{Weight Regularization}
For our weight regularization-based method, we employ the regularization term used in EWC \cite{Kirkpatrick3521}. Thus, we compute the diagonal of the Fisher information matrix $F_i$ at the beginning of each episode and compute the objective as below:

\begin{equation}
    \theta_{t}^{*} = \operatorname*{arg\,min}_\theta [{\mathscr{L}_{TTS}(D_{tr}^t|\theta=\theta_{t-1}^*)} + \sum_{i}\frac{\lambda}{2} F_{i}(\theta_i - \theta_{t-1,i}^*)]
\end{equation}

where $\theta_{t-1, i}$ indicates the optimal weights of the previous episode for parameter $i$ of the model, and $F_i$ determines how important parameter $i$ is to the previous tasks. $\lambda$ is a hyperparameter that controls the importance of the older tasks. 

\subsubsection{Experience Replay (ER)}
In Experience Replay, we keep a buffer of samples from previous speakers and combine them with the data from the current task for the rehearsal mechanism in every episode. The combination of buffer items and the current dataset can be done in several ways. For example, one can first train the model on the data from the current speaker in every training epoch and then compute the gradient for the buffer items and optimize for them at the end of the epoch. Another possibility would be running the replay phase after a particular number of training epochs for the current speaker. In our experiments, we recognized that mixing buffer with the main dataset plus shuffling works the best. This is probably due to the fact that mixing the buffer items with the main dataset increases the possibility of experience replay in the middle iterations of every epoch, and it prevents the model optimization from moving in a speaker-specific solution and acts as a regularizer.

\SetKwInput{KwInput}{Input}                
\SetKwInput{KwOutput}{Output}

\newcommand{\Model}{\operatorname{Model}}
\newcommand{\algtext}{\operatorname{text}}
\newcommand{\algmel}{\operatorname{mel}}
\newcommand{\Optimize}{\operatorname{Optimize}}

\begin{algorithm}

\SetAlgoLined
\KwInput{$\mathcal{D}_{tr}^{t}$, $\mathcal{D}_{ev}^{t}$, Buffer $B_{t-1}$, $\theta_{t-1}^*$, Buffer Size $N_B$}
\KwOutput{$\theta_{t}^*$, $B_{t}$}
    
 Initialize Model with $\theta_{t-1}^*$\;
 \While{Early stopping criterion is not met}{
  $\theta_{t}^* \xleftarrow{} \Optimize(\Model; {D}_{tr}^{t} \cup B_{t} )$\;
  
 }

 \eIf{ER-KD}{
    $B_{t}^{'} = \{\algtext^k; k=1,...,N_B\} \sim D_{tr}$\;
    $B_{t}^{KD} = \{(t, \algmel=\Model(t| \theta_{t}^*); \forall t \in B_{t}^{'} \} $\;
    $B_{t} \xleftarrow{} B_{t-1} \cup B_{t}^{KD} $\;
   }{
    $B_{t}^{'} = \{(\algtext^k, \algmel^k); k=1,...,N_B\} \sim D_{tr}$\;
    $B_{t} \xleftarrow{} B_{t-1} \cup B_{t}^{'} $\;
  }
  
 \caption{Experience Replay for Speaker $S_t$}
 \label{alg:ER}
 \vspace{-0.5cm}
\end{algorithm}

We sample the buffer items randomly, and in our experiments, we keep the same number of samples for every previous speaker. In addition to directly storing the batch items in the buffer, for Experience Replay with Knowledge Distillation (ER-KD), we keep the input transcript, and the output Mel-spectrogram obtained from the best-trained model for all past speakers. In Algorithm \ref{alg:ER} we explain the steps of ER and ER-KD.

\subsubsection{Experience Replay with Buffer Replication (ER-BR)}
Since the performance of ER-based methods depends on the buffer size, limiting buffer size has a negative impact on speech diversity. To evaluate the accuracy of a very limited case, we keep only one sample per speaker in the buffer. We found online replication of buffer elements more effective. To do so, we replicate the buffer items $K$ times, with $K$ being the replication factor. In general, the results were more stable in terms of the pace of speech.

\section{Results}
We perform all experiments on three random order of sequence $S=\{s_1, s_2, ..., s_m \}$ of speakers by shuffling them to ensure that the results are not only dependent on the order. Some audio samples are available on the demo page \footnote{\url{https://d872c.github.io/Interspeech21_Demo}}.

\begin{figure}[t!]
    \centering
    \includegraphics[width=0.49\linewidth]{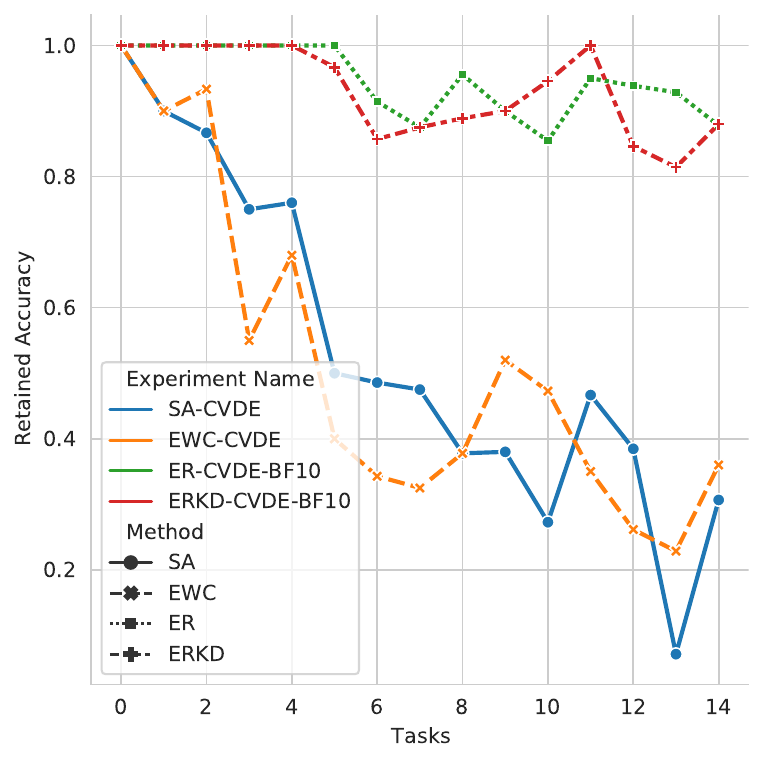}
    \includegraphics[width=0.49\linewidth]{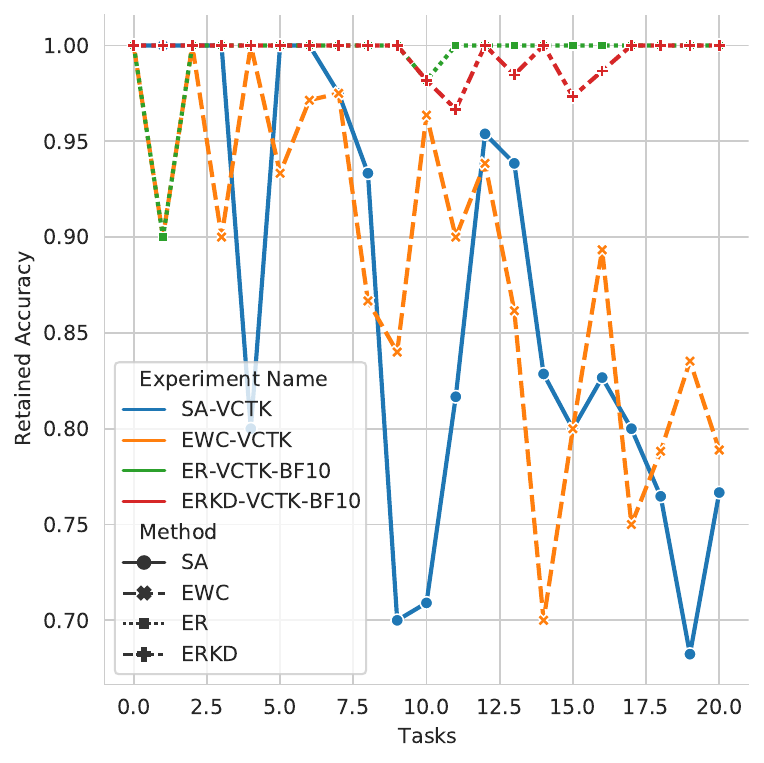}
    \caption{Comparison of RA in SA, EWC, ER and ERKD methods for CVDE(lefT) and VCTK(right) dataset.}
    \label{fig:ra_comp}
    \vspace{-0.5cm}
\end{figure}

\begin{figure*}[t!]
    \begin{subfigure}[t]{0.5\textwidth}
        \includegraphics[width=0.24\linewidth]{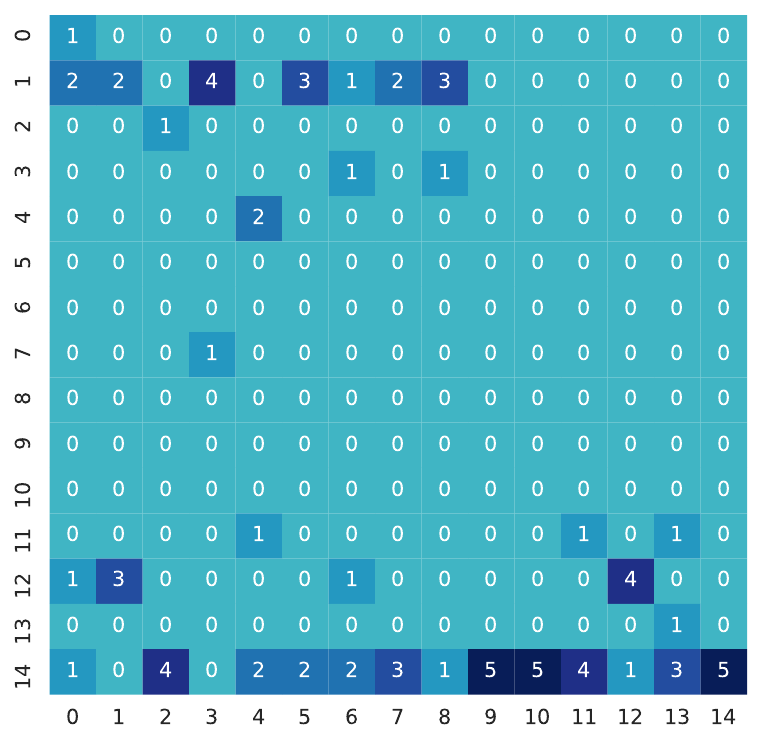}
        \includegraphics[width=0.24\linewidth]{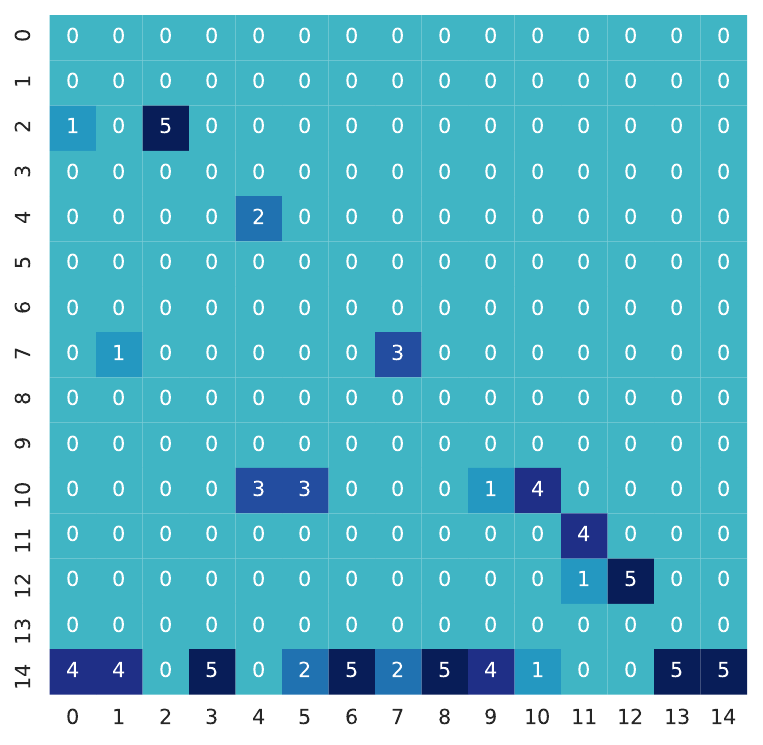}
        \includegraphics[width=0.24\linewidth]{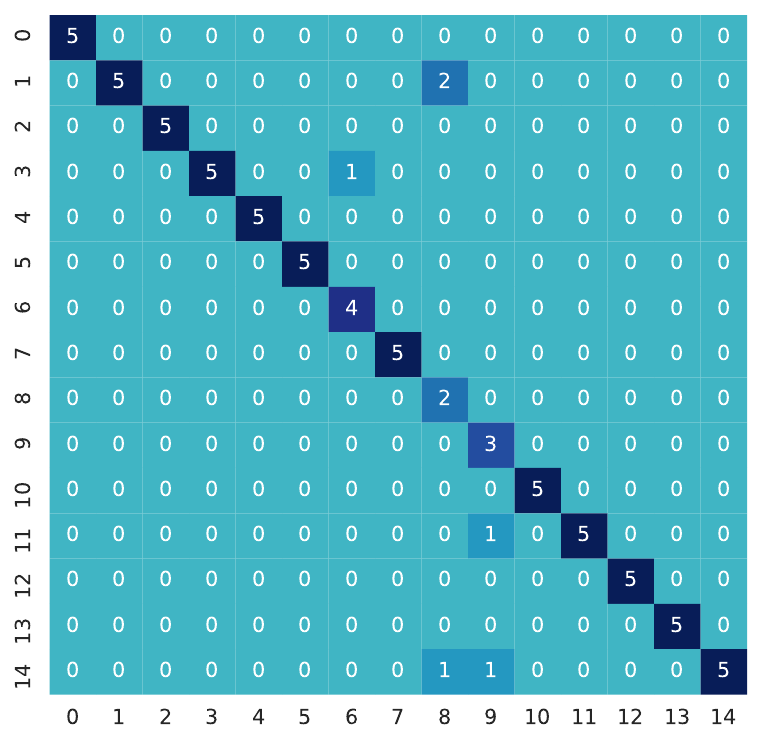}
        \includegraphics[width=0.24\linewidth]{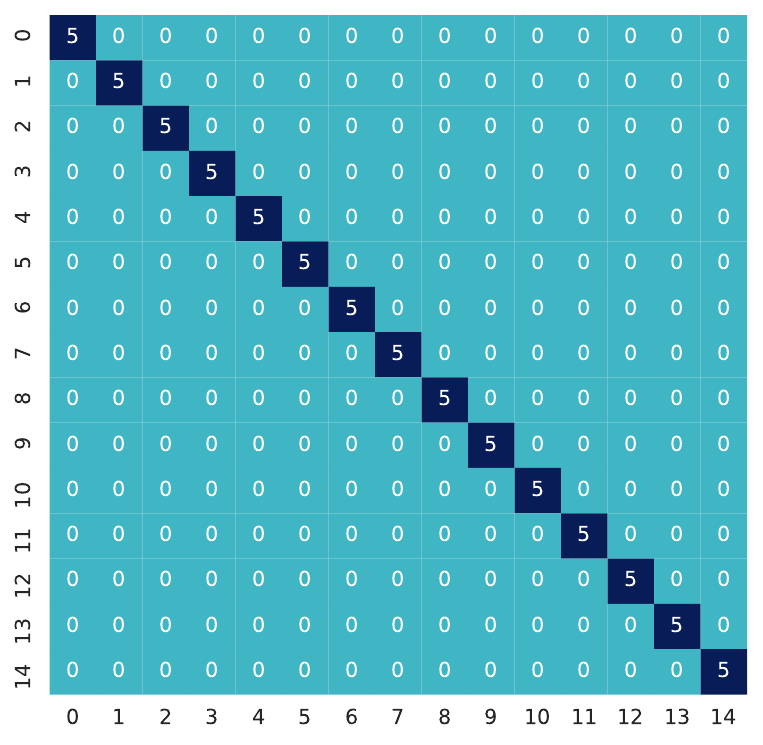}
        
        \caption{Confusion matrix for SA, EWC, ER and JT methods respectively from left to right. Each column/row corresponds to one speaker with same order as in the sequence of speakers in the training episodes.}
    \end{subfigure} 
    \hspace{2mm}
    \begin{subfigure}[t]{0.5\textwidth}
        \includegraphics[width=0.24\linewidth]{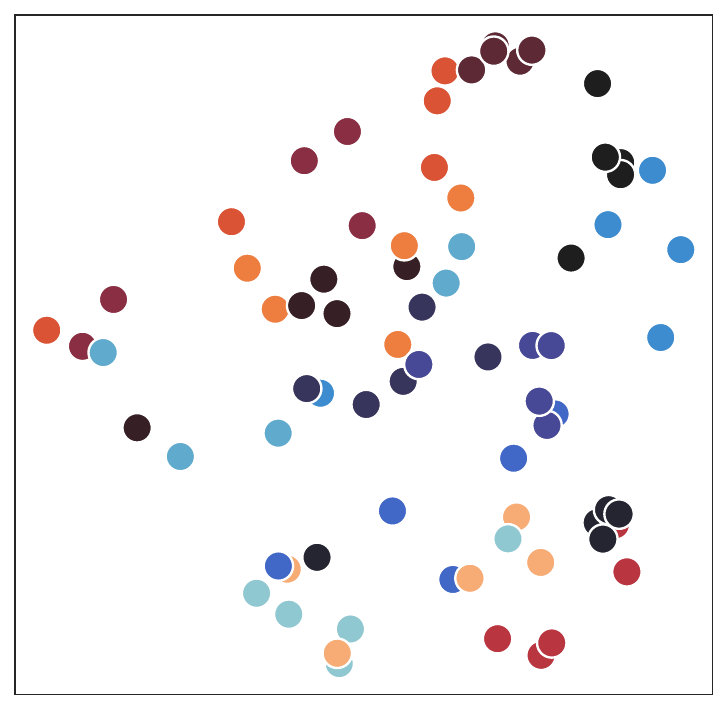}
        \includegraphics[width=0.24\linewidth]{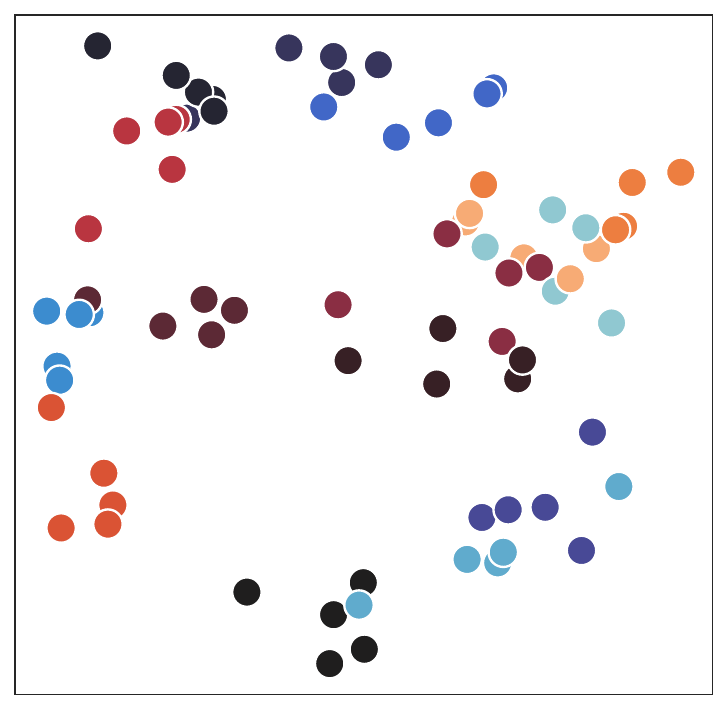}
        \includegraphics[width=0.24\linewidth]{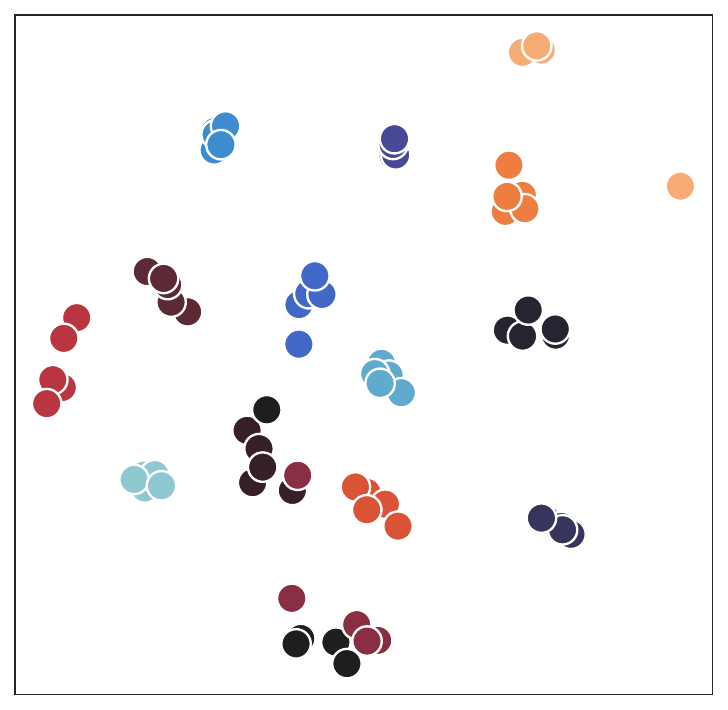}
        \includegraphics[width=0.24\linewidth]{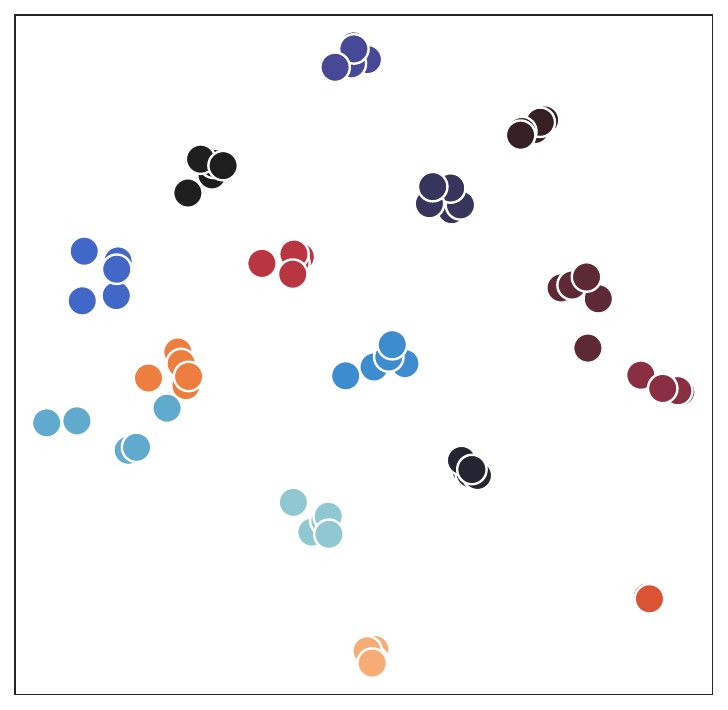}

        \caption{t-SNE visualization of speech embeddings computed from 5 different utterances synthesized for each speaker with the methods SA, EWC, ER, and JT, respectively, from left to right. Each color represents a speaker.}
    \end{subfigure}
    \caption{Visualization of accuracy results (left) and speech embedding visualization (right) for the final step (episode) of three different methods with the same random speaker order. In JT, by final episode we mean the only episode of training which consists of all speakers.}
    \label{fig:cf_tsne}
\end{figure*}

\subsection{Comparison of Different Methods}
In Figure \ref{fig:ra_comp} we compare the RA over one random sequence of episodes for both CVDE and VCTK datasets. Despite the better stability of VCTK in SA, which is mainly caused by the richer initialization $\theta_{init}^*$ obtained from its larger PR-Split, both SA and EWC methods are prone to forgetting. In general, we noticed that ER-based methods are much more stable, and EWC performed only a little bit better than SA in some cases. In ER method, it is expected to get better results by increasing the buffer size, but the goal is to keep the number of samples low such that the training time for new speakers remains almost constant over time. As presented in Table \ref{tab:er_buffersize} we saw that keeping ten audio samples per speaker gives reasonably good results.

To get insights into the details of forgetting, we visualize the confusion matrix of speaker verification results and t-SNE reduction of speech embeddings of the synthesized speech waveforms in the final episode of one of the settings in Figure \ref{fig:cf_tsne}. It is evident that most of the synthesized speech files are classified as the last speaker in the sequence in SA and EWC. Unlike EWC and SA, ER generates speech waveforms that are more diverse, and most of the older speakers are correctly classified. This implicitly shows that the model has not forgotten how to generate speech waveforms for the older speakers given their speaker identifiers (embeddings). 

\vspace{-.2cm}

\begin{figure}[t]
  \centering
\includegraphics[width=0.49\linewidth]{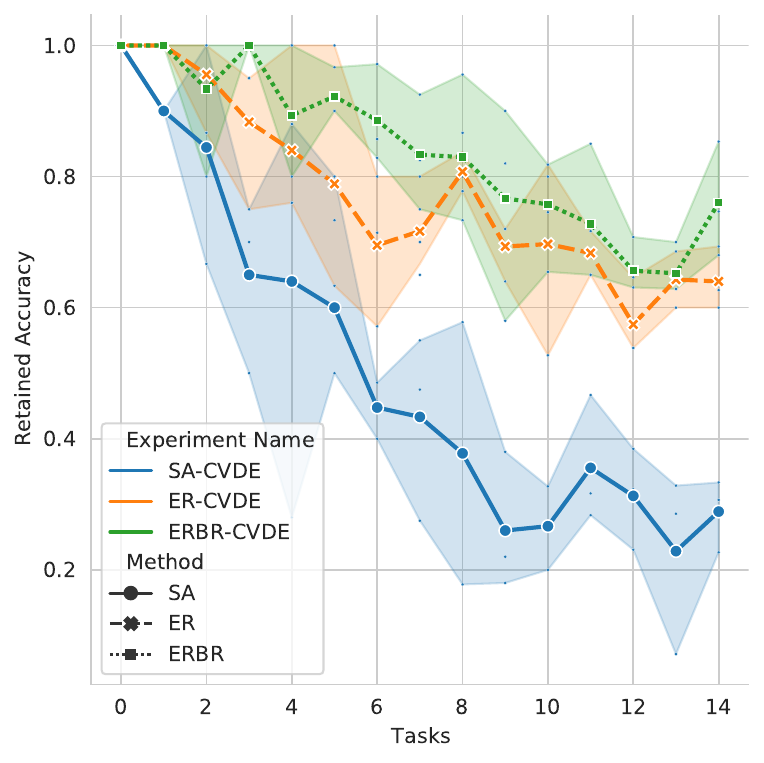}
\includegraphics[width=0.49\linewidth]{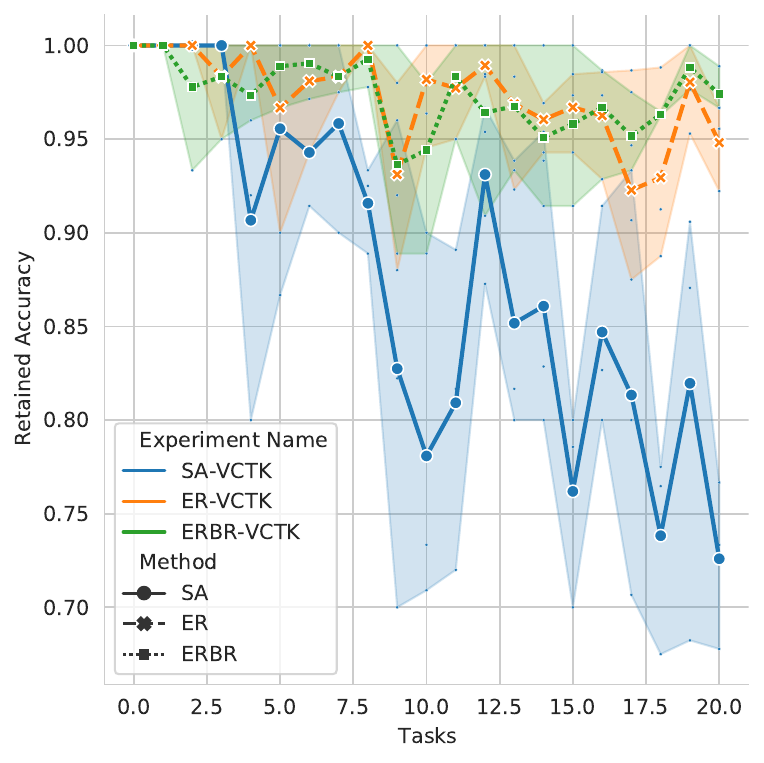}

  \caption{Plots showing higher retained accuracy and robustness with confidence interval obtained by ER-BR compared to ER and SA for three distinct random orders of speaker sequences for CVDE (left) and VCTK (right) datasets.}
  \label{fig:ra_erbr}
  \vspace{-0.5cm}
\end{figure}

\begin{table}[th]
  \caption{Table shows per-speaker accuracy for the first three speakers in one of the speaker sequences of CVDE dataset for ER method with buffer size 5 when the model is adapted to the 11th-15th speakers.}
  \label{tab:per_speaker_ra}
  \centering
\begin{tabular}{ c|c|c|c|c|c  }
\hline
\textbf{Speaker} & $S_{11}$ & $S_{12}$ & $S_{13}$ & $S_{14}$ & $S_{15}$ \\
\hline
 $S_1$ & 0.8 & 1.0 $\uparrow$ & 0.8 $\downarrow$ &  1.0 $\uparrow$ &  0.2 $\downarrow$ \\
 $S_2$ & 1.0 & 1.0 & 1.0 &  1.0 &  0.4 $\downarrow$\\
 $S_3$ & 0.8 & 0.6 $\downarrow$ & 1.0 $\uparrow$ &  1.0 &  0.4 $\downarrow$\\
 \hline

\end{tabular}
\end{table}
 
 \vspace{-.3cm}
\subsection{Backward Transfer}
When adapting to a new speaker, its impact on each previous speaker can be different. In our setup, backward transfer \cite{riemer2018learning} happens when new speaker's training increases another speaker's verification accuracy. In Table \ref{tab:per_speaker_ra} we show an example of positive and negative backward transfer by up- and down- arrows respectively. The same effect is also visible in the Figures \ref{fig:ra_comp} and \ref{fig:ra_erbr}. 
\begin{table}[th]
  \caption{Effect of buffer size on the forgetting (right columns) and retained accuracy (left columns) of the last 3 episodes}
  \label{tab:er_buffersize}
  \centering
\begin{tabular}{ c|c c|c c|c c  }
\hline
\textbf{Experiment} & \multicolumn{2}{c|}{$S_{T-2}$} & \multicolumn{2}{|c|}{$S_{T-1}$} & \multicolumn{2}{|c}{$S_{T}$} \\
\hline
 CVDE-BS:1 & 0.54 & 0.48 & 0.60  & 0.42 & 0.62 & 0.39   \\
 CVDE-BS:2 & 0.68 & 0.32 & 0.73  & 0.26 & 0.81 & 0.17   \\
 CVDE-BS:5 & 0.78 & 0.22 & 0.86  & 0.08 & 0.84 & 0.10   \\
 CVDE-BS:10 & \textbf{0.94} & 0.03 & \textbf{0.93}  & 0.00 & 0.88 & 0.04   \\
 CVDE-BS:20 & \textbf{0.94} & 0.03 & \textbf{0.93}  & 0.00 & \textbf{1.0} & -0.07   \\
 \hline
 
 VCTK-BS:1 & 0.99 & 0.01 & \textbf{1.0}  & 0.00 & 0.95 & 0.05   \\
 VCTK-BS:2 & 0.98 & 0.03 & 0.98  & 0.03 & 0.97 & 0.04   \\
 VCTK-BS:5 & 0.96 & 0.04 & \textbf{1.0}  & 0.00 & 0.99 & 0.01   \\
 VCTK-BS:10 & \textbf{1.0} & 0.00 & \textbf{1.0}  & 0.00 & \textbf{1.0} & 0.00   \\
 VCTK-BS:20 & \textbf{1.0} & 0.00 & \textbf{1.0}  & 0.00 & 0.99 & 0.01   \\
 \hline

\end{tabular}
\end{table}

\vspace{-0.5cm}
\subsection{ER-BR Results}
In Figure \ref{fig:ra_erbr} we visualize the results for the RA of ER-BR method with confidence interval for buffer size 1 and replication factor of 10 acquired by 3 distinct order of speaker sequences.  The improvement in RA is especially important for the speakers seen at the beginning of the sequence. In Table \ref{tab:er_vs_erbr} the increase in the RA of $S_1$ and $S_2$ shows that distribution of buffer sample replica in the training batches of each episode can help with preserving old speakers more effectively in extreme buffer limitation setups.

\begin{table}[th]
  \caption{Per-speaker accuracy for the first three speakers after the final episode of training for ER with buffer size 1 and ER-BR with replication factor of 10.}
  \label{tab:er_vs_erbr}
  \centering
\begin{tabular}{ c|c c c }
\hline
\textbf{Speaker} & $S_1$ & $S_2$ & $S_3$ \\
 \hline
 ER - CVDE & 0.20 & 1.0 & 0.20 \\
 ER-BR - CVDE & \textbf{1.0} & 1.0  & \textbf{0.60} \\
 \hline
\end{tabular}
\end{table}

\vspace{-0.2cm}
\section{Conclusion}

We proposed a framework for the continual adaptation of speakers in TTS systems. To measure the performance of continual adaptation in terms of speech diversity, we employed a linear speaker verifier for speech embeddings extracted by a pre-trained speech encoder. Although this approach does not necessarily evaluate all aspects of TTS model evaluation, like speech quality, it can be considered as a good approximation of model performance and sample diversity. We also demonstrated that experience replay could be a very effective method for continual learning in TTS and can prevent catastrophic forgetting to a great extent compared to sequential fine-tuning and EWC.

\nocite{NEURIPS2019_9015}

\bibliographystyle{IEEEtran}

\bibliography{mybib}


\end{document}